\documentclass[a4paper,10pt]{article}
\usepackage[utf8]{inputenc}
\usepackage{graphicx}
\usepackage{indentfirst}

\begin{document}

\title{Preliminary Wildfire Detection: A Dataset and Challenges}
\author{Samarth Shah \\\\
2021 Summer STEM Institute}
\date{\today}
\maketitle

\begin{abstract}
Wildfires are uncontrolled fires in the environment that can be caused by humans or nature. In 2020 alone, wildfires in California have burned 4.2 million acres, damaged 10,500 buildings/structures, and killed more than 31 people, exacerbated by climate change and a rise in average global temperatures \cite{insurance_information_institute} \cite{cragcrest_2018}. This also means there has been an increase in the costs of extinguishing these treacherous wildfires. The objective of the research is to detect forest fires in their earlier stages to prevent them from spreading, prevent them from causing damage to a variety of things, and most importantly, reduce or eliminate the chances of someone dying from a wildfire. A fire detection system should be efficient and accurate with respect to extinguishing wildfires in their earlier stages to prevent the spread of them along with their consequences. Computer Vision is potentially a more reliable, fast, and widespread method we need. The current research in the field of preliminary fire detection has several problems related to unrepresentative data being used to train models and their existing varied amounts of label imbalance in the classes (commonly fire and non-fire) of their dataset. We propose a more representative and evenly distributed data through better settings, lighting, atmospheres, etc., and class distribution in the entire dataset. After thoroughly examining the results of this research, it can be inferred that they supported the dataset’s strengths by being a viable resource when tested in the real world on unfamiliar data. This is evident since as the model trains on the dataset, it is able to generalize on it, hence confirming this is a viable Machine Learning setting that has practical impact.

\end{abstract}

\newpage

\section{Introduction}

Wildfires are uncontrolled fires in the environment that can be caused by humans or nature. These fires are dangerous to both humans and nature as they have the potential to destroy a copious amount of property and also damage the ecosystem. In 2020 alone, wildfires in California have burned 4.2 million acres, damaged 10,500 buildings/structures, and killed more than 31 people \cite{insurance_information_institute}. The solution must allow for endangering wildfires to be found before they can spread which can save hundreds of thousands of dollars typically spent on extinguishing fires, repairing damaged property, and most importantly, it can save lives. Although there are many applications of Machine Learning in the field of early wildfire detection, many are too costly to implement and/or not feasible to work with. 
\begin{figure}[htp]
    \centering
    \includegraphics[width=3.5cm]{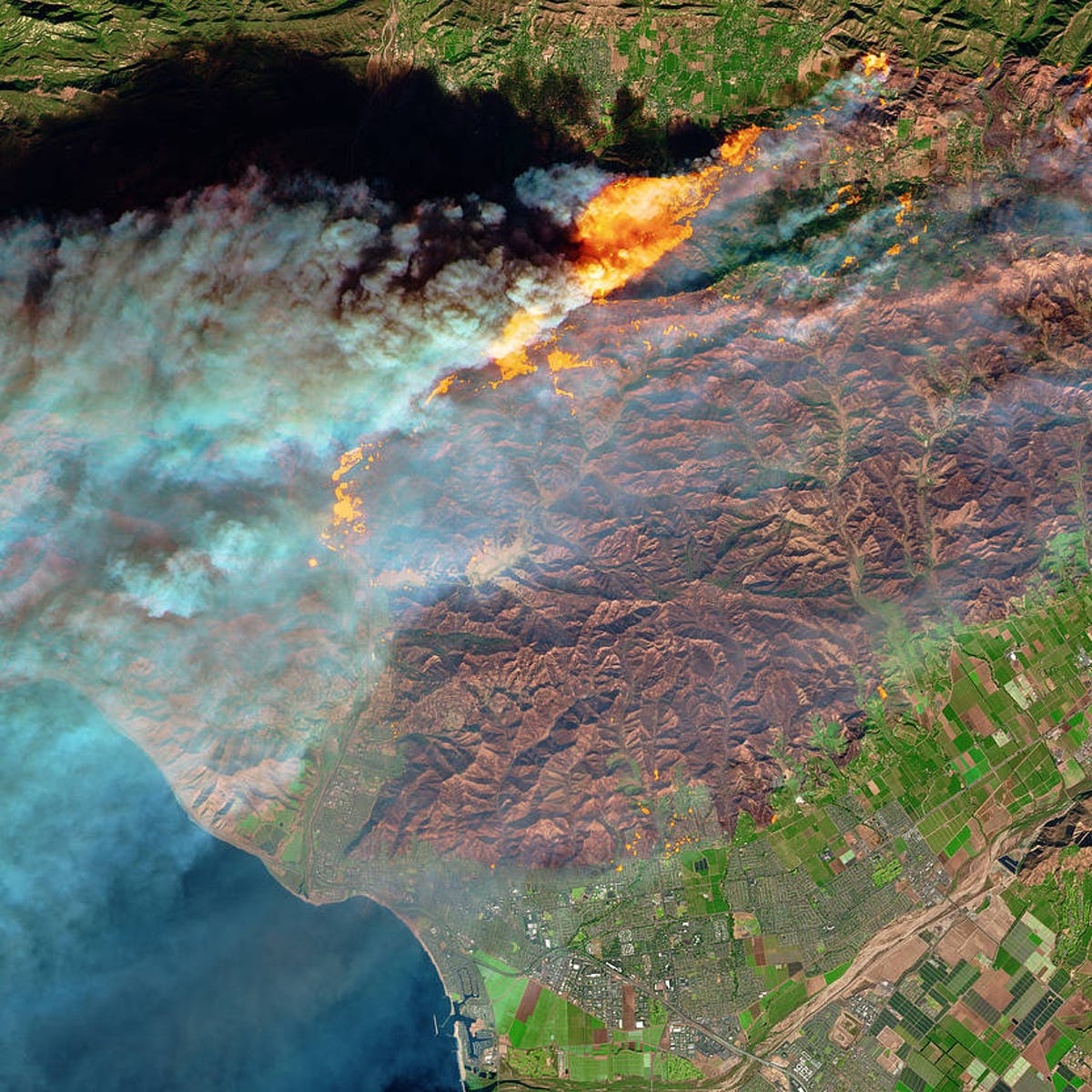}
    \includegraphics[width=3.5cm]{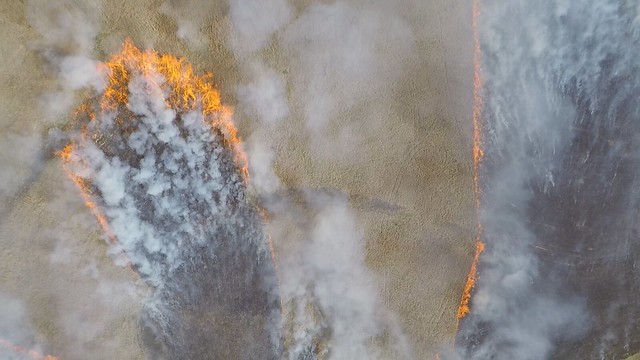}
    \includegraphics[width=3.5cm]{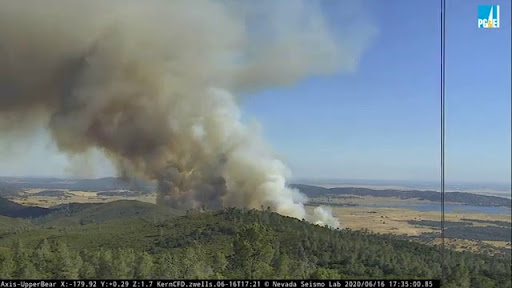}
    \caption{Left: An Image Captured Using A Satellite-based Solution \cite{hub_2021}. This solution is more costly and does not detect wildfires in their earlier stages. Middle: An Image Captured Using An Aerial-based Solution \cite{elliott_voican_singh_2021}. This solution is too costly and not very feasible due to operating expenses and short flight time. Right: An Image Captured Using A Terrestrial-based Solution \cite{alertwildfire_faculty}. This solution is relatively more practical since it is cheaper and more feasible, and can cover large areas of land.}
    \label{fig:gate}
\end{figure}
Computer Vision is a necessary tool as part of a larger wildfire solution. Computer Vision allows for a more efficient and cheap solution to a multitude of applications as it can repetitively carry out tasks at a constant faster pace with less human error \cite{simplilearn_2021}. They are able to automate tasks that wouldn't be feasible for humans to complete at a constant pace on a regular basis. ALERTWildfire is a large scale PTZ camera network used to monitor landscapes. Currently, the organization has given humans the responsibility of catching wildfires during their preliminary stages. Computer Vision can allow for an improved solution to wildfires by automating the human-dependent process being carried out today. This Computer Vision solution can be easily expanded and adopted by others. With ALERTWildfire’s large-scale camera network, I plan to use their camera data as they are able to look at all of the land on the western part of The United States that have a risk of sustaining wildfires. ALERTWildfire is an organization that partners with several other groups stationed in the western part of The United States. By doing so, they’re able to provide such a network of cameras in which each partner holds a stake. In the future, this camera network will be expanded globally to places all over the world that are susceptible to contain wildfires. These terrestrial-based solutions make way for an Image Recognition algorithm that can constantly scan for fires in multiple locations that are likely to contain wildfires with wide-angle pan-tilt-zoom cameras \cite{alertwildfire_faculty}.
\begin{figure}[t]
    \centering
    \includegraphics[width=6cm]{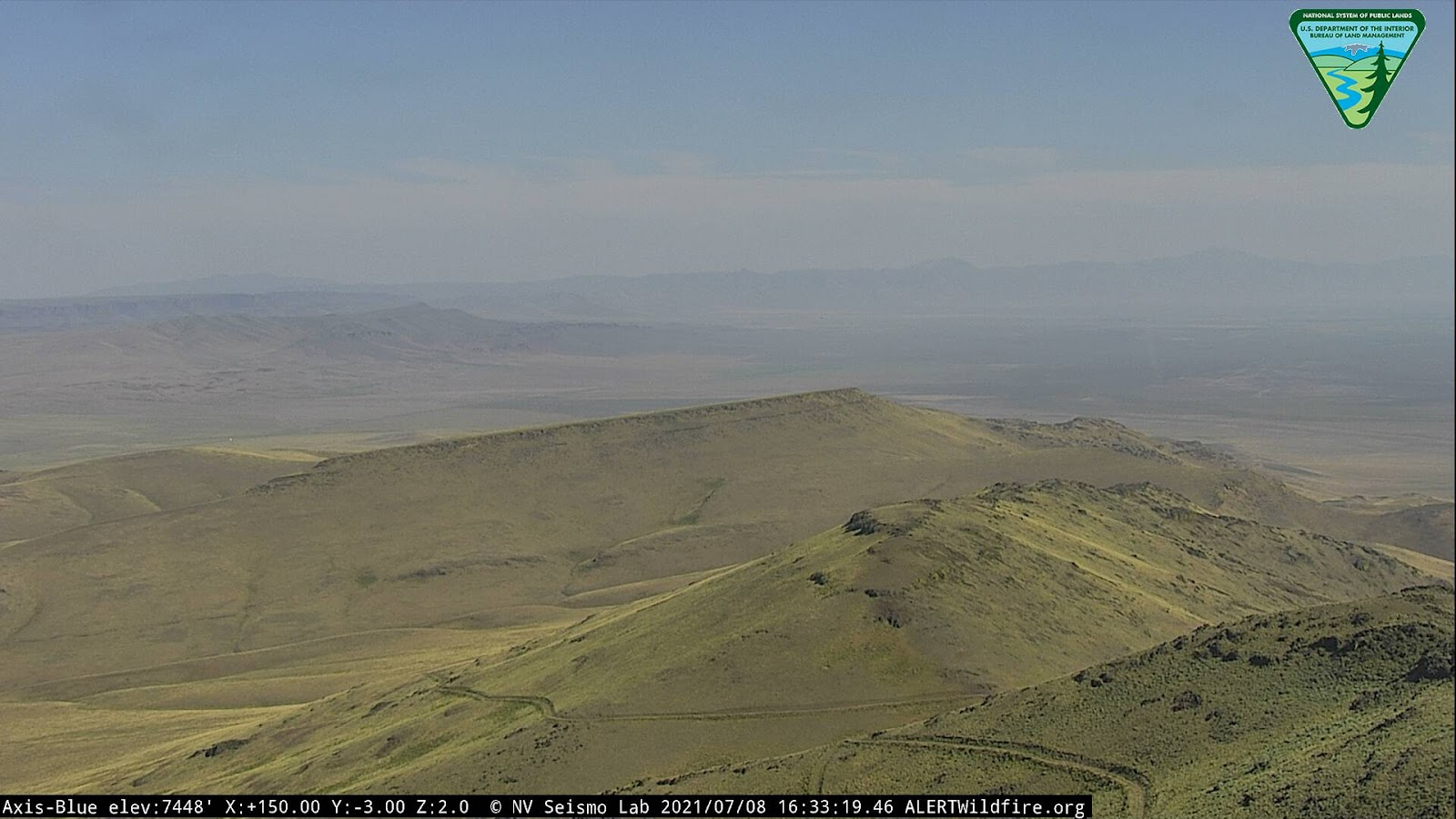}
    \includegraphics[width=6cm]{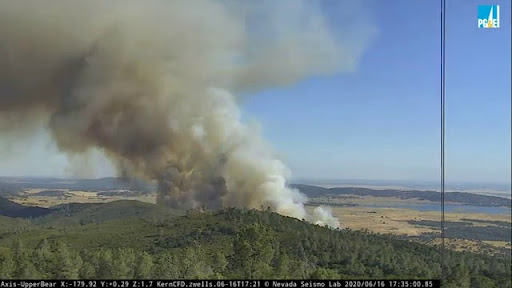}
    \caption{Left: Landscape Without Fire, Right: Landscape With Fire. Both images originate from a wide-angle pan-tilt-zoom camera view provided by ALERTWildfire. Fires usually happen in the day, and smoke is the biggest visual characteristic for detection. Evident through comparing each image to the other, the image on the left does not have a fire despite their being smoke in the air above the landscape since it is not sprouting from the land. In contrast, the image on the right does have a fire since there is smoke coming from the land. Although there are clouds in the background that could be classified as smoke, they are not considered as smoke caused by a wildfire as they are not coming from the ground.}
    \label{fig:gate}
\end{figure}
On this topic, there are a few other research papers that are based on the same theme of using  Machine Learning and Computer Vision technology to detect fires on their earlier stages to mitigate the damage they produce \cite{DBLP:conf/eann/SharmaGGF17} \cite{DBLP:journals/mta/LuoZLH18}. However, these papers have several issues and limitations regarding practicality due to their failure to consider the Machine Learning obstacles. This is a very common issue found in these papers simply because their respective authors tend to ignore the implications of the form of the dataset they present to their models. Although the accuracies presented by these papers may seem enticing at first glance, they are not valid representations of how well their models would perform when given real-world data from common cameras used for early wildfire detection, such as the ones provided by ALERTWildfire. For instance, one common issue that circulates through the research in this field is the issue of Model Robustness–the model’s ability to identify the important features that distinguish one class from another in the case of image recognition. This is due to the several papers in the same field that have models unable to pick up on distinguishing factors to accurately determine whether or not a given image contains a wildfire. Therefore, this would result in a not very robust model as the data used for training is dissimilar to what is  seen when the model tests on real-world data. Consequently, this problem can lead to an issue with Calibration–the model’s confidence or the probability that its predictions are reliable. If the model is not very robust, it could negatively affect its Calibration as it would decrease the probability that its predictions are reliable when applied to a real-world scenario. The research being carried out will also focus on mitigating these prevalent issues by using more representative and evenly distributed data in terms of the image’s setting, lighting, atmosphere, etc., and class distribution in the entire dataset. The data was also masked on the parts that were not correlated with the determination of whether or not there exists a fire in a given image. The data was split such that 75\% of it was allocated for training purposes and 25\% of it was allocated for purposes testing the accuracy. There were 4000 images in which 2000 of them were images of fires and 2000 of them were images of landscapes. By doing these things, I will resolve or mitigate the issues evident in neighboring research papers. For instance, keeping the dataset with a more representative setting, lighting, atmosphere, etc. will allow me to train the model on what it will be expected to input which will increase the validity of the accuracy. By masking parts of the images in the dataset, I will also lessen the bias present in the images to also increase the validity of the accuracy. Through carrying out the experimentation phase of this research and closely observing the results, one can note the dataset's attempt to mitigate the bias found in neighboring papers through its distinguishing features mentioned above. A Convolutional Neural Network was able to train on this dataset, lessen its loss, and increase its accuracy.

\section{Literature Review}

I have determined some of the practical problems faced by researchers across the board regarding preliminary wildfire detection by looking through several pieces of literature relating to real-world applications using Artificial Intelligence and collecting data that will be used to train the model. I also noted common techniques used in the algorithms and general, recurring problems that show up with the image recognition implemented for early wildfire detection. Apart from Computer Vision, there are other solutions referred to by some in the field of preliminary wildfire detection. One of the most commonly-referred alternatives to using Image Recognition is to implement terrestrial-based sensors that are stationed at spots likely to contain wildfires \cite{DBLP:journals/sensors/BarmpoutisPDG20}. Some also look at aerial-based solutions using systems such as drone networks \cite{DBLP:journals/sensors/BarmpoutisPDG20}. The issue with both of these approaches is that they are too costly and not efficient enough to be deployed on a large scale, which is necessary when there is not one designated location and designated period of time where and when wildfires ignite. I have included illustrations and diagrams of some of the more prevalent problems that occur in Computer Vision research.

\textbf{Model Robustness} refers to the model’s ability to identify the important features that distinguish one class from another in the case of image recognition. Oftentimes, data used by researchers would contain images of fires that were very similar to each other and would often be located on the same spot of the image, would have the same hue of the flame, etc. For instance, if an algorithm was trained on images that were pointed at the same location/landscape with the same features, it could have only learned how to detect wildfires in that particular setting and for that reason would not be a very robust model as it would not be using the features that distinguish an image from having a wildfire as opposed to not having one. Not having a robust model in the field of wildfire detection would decrease the representativeness of the displayed accuracy of the model if it refers to its accuracy as its ability to detect wildfires in general \cite{DBLP:journals/remotesensing/GovilWBP20}. In turn, this would result in a decrease in validity of the accuracy showcased by several of the research papers–in the field of preliminary wildfire detection–that have datasets filled with skewed images that had several other indicators of fires such as the orientation of the image, the setting of the image, etc. \cite{beneducealertwildfire}

\begin{figure}[htp]
    \centering
    \includegraphics[width=6cm]{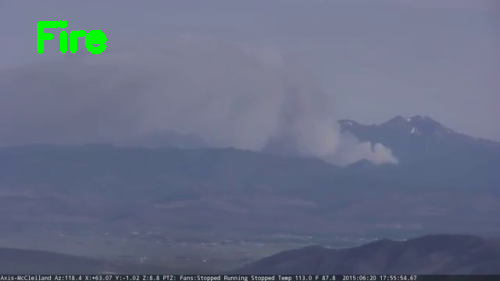}
    \includegraphics[width=6cm]{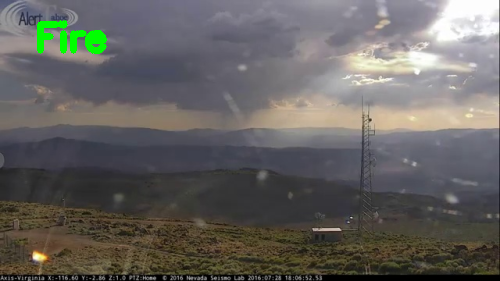}
    \caption{Left: Detected Wildfire, Has Wildfire. Right: Detected Wildfire, No Wildfire. An Illustration of The Model Robustness Problem – The Model’s Tendency To Perform Poorly From “Subtle Differences” In The Input Image. Although the left image is a correct prediction as there is a wildfire in the image, the right image is an incorrect prediction as there is no wildfire in that image. This is due to an issue with the Model Robustness since a subtle change in the image resulted in a false prediction. The gray areas of the smoke, which are often used to determine the existence of a wildfire, look relatively similar and thus have not changed much between the two images.}
    \label{fig:gate}
\end{figure}

\textbf{Calibration} refers to the model’s confidence or the probability that its predictions are reliable. Calibration is especially important in the field of early wildfire detection as any individual in the field would want to ensure that they catch all the wildfires that occur without missing any, but at the same time, not wanting to overtrain the model to the point where it would detect false positives \cite{DBLP:journals/corr/abs-2003-00646}. 

\textbf{Generalization} refers to the ability of a trained model to perform predictions on unseen/unfamiliar data. As touched on earlier regarding the model’s robustness, researchers would often find their models not performing at the accuracy listed when running on unfamiliar data proving their models haven’t completely generalized on all wildfires \cite{DBLP:conf/eann/SharmaGGF17}. This relates to the robustness of the model as discussed earlier since a model that is less robust tends to not be as generalizable. This means generalization and model robustness has a direct relationship. Relating the previously brought up skewed image with the topic of generalization, if the model trains on images that have odd orientations, unrepresentative settings, and other sources of dataset bias, it will not generalize its weights on realistic data that it will input when applied to cameras in an outdoor setting where practically all wildfires occur \cite{beneducealertwildfire}. Moreover, as a model continues to train on familiar data, it increases its confidence in it and this will result in increased poor performance on unfamiliar data. This phenomenon is typically referred to as a Domain Shift \cite{DBLP:conf/cvpr/Sankaranarayanan18}.

\begin{figure}[htp]
    \centering
    \includegraphics[width=9cm]{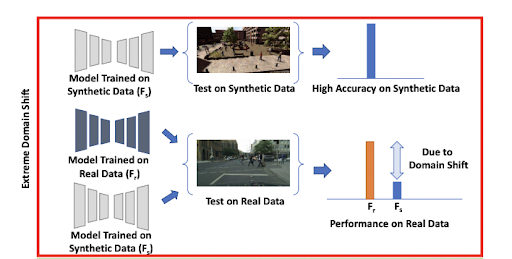}
    \caption{A Visualization of The Domain Shift Problem \cite{DBLP:conf/cvpr/Sankaranarayanan18}. Evident from the diagram pictured, there is a decrease in performance on real data in comparison to the model's performance on synthetic data. This can be attributed to the Domain Shift phenomenon.}
    \label{fig:gate}
\end{figure}

\textbf{Label Imbalance} refers to the problem faced by a model regarding predictions due to the number of examples for each class in the training set being unbalanced. In the realm of wildfire detection and mitigation, Label Imbalance must be taken into account as there is always more data with images of without wildfires than there are with wildfires. Label Imbalance is difficult because models will tend to output majority class predictions and will lean in opposition to the minority class. One way to mitigate the unwanted effects of Label Imbalance is by implementing a Class Rectification Loss Function to automatically allow for the model to correct for the “dominant effect” of the majority classes through observing minority classes and their sampled boundaries \cite{DBLP:journals/pami/DongGZ19}. Through reviewing the literature in the same field of early wildfire detection, Label Imbalance is evident in several papers through datasets having relatively more data in the class with wildfires. Due to this, their models tend to be biased towards detecting wildfires more easily as they are fine-tuned on more examples of wildfires than images without wildfires \cite{DBLP:conf/eann/SharmaGGF17}.

\begin{figure}[htp]
    \centering
    \includegraphics[width=9cm]{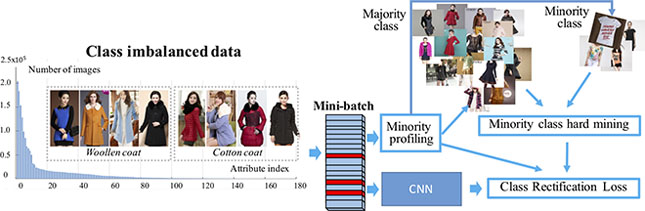}
    \caption{An Illustration of A Label Imbalance and The Involvement of Class Rectification Loss \cite{DBLP:journals/pami/DongGZ19}. Having more data in one class would introduce a bias in the model's judgement of data towards the majority class.}
    \label{fig:gate}
\end{figure}

\section{Research Questions}

\begin{enumerate}
    \item How is the dataset collected and why is it comprehensive and inclusive of wildfire situations?
    \item Why is the dataset able to mitigate common computer vision problems such as different atmospheres, model robustness, etc. and what makes it well-suited for improving not only the model’s accuracy, but also its validity?
    \item What makes this dataset difficult?
\end{enumerate}

\section{Methodology}

The dataset was consists of 2000 images of wildfires on landscapes susceptible to containing wildfires and 2000 images of landscapes that are susceptible to containing wildfires but don’t contain any wildfires at the instance they were taken. The images of wildfires were collected through website scraping methods using PyTube, a Python module that can scrape videos from a YouTube channel and convert them to MP4 files, by using ALERTWildfire’s YouTube channel which contains hundreds of videos of fires that occurred on landscapes susceptible to containing wildfires \cite{alertwildfire_youtube} \cite{pytube}. Once the videos were scraped and converted into MP4 files, OpenCV, a Python module that has a plethora of methods that deal with images, was used to capture frame from the video every 10 seconds \cite{opencv}. By doing this, the dataset is ultimately able to incorporate more variety in the images of wildfires as it contains various angles of them while also maintaining frames that contained wildfires that are representative of what a real-world input for the model would look like. The images of landscapes without wildfires were collected through website scraping methods using Beautiful Soup and PyAutoGUI, two python libraries that are able to automate website functions such as searching on the website and selecting buttons and attributes on the website page. ALERTWildfire’s website page was scraped using these modules by saving each image into the appropriate section of the dataset by having Beautiful Soup direct the algorithm to the right website page representing each of the cameras provided by the organization \cite{beautiful_soup_documentation}. PyAutoGUI was used to automatically operate the functions on the site to save each of the captured images from the cameras \cite{pyautogui_documentation}. By doing this, the dataset is also able to incorporate a larger range of data regarding images of landscapes susceptible to containing wildfires. All in all, what makes this dataset so comprehensive and inclusive of wildfire situations is that it uses images of landscapes that have wildfires or are susceptible to containing wildfires. Therefore, when applied to a real-case scenario on unfamiliar data, a model trained on this dataset can be more robust and will hence be able to generalize on more applicable data. As mentioned earlier, this dataset has been collected through images that are very representative of what the algorithm would capture as an input in the real world. In addition to this, the images were also masked on certain parts that would cause bias and did not correlate with a higher or lower chance of there being a wildfire in those images. Both of these things definitely help mitigate the bias and unrepresentativeness present in datasets that are typically showcased in neighboring research papers in the same field of preliminary wildfire detection. These neighboring research papers, as mentioned in other segments of this paper, have issues regarding Model Robustness and Generalization which often occur from the datasets presented in them that often have a bias due to the unrepresentative setting, atmosphere, lighting, etc. The dataset formed from this research mitigates and in some cases solves problems such as the ones listed above as the masks eliminate the possibility of non-wildfire-related factors interfering with the model’s weights during training.

\section{Results and Discussion}
\begin{figure}[htp]
    \centering
    \includegraphics[width=9cm]{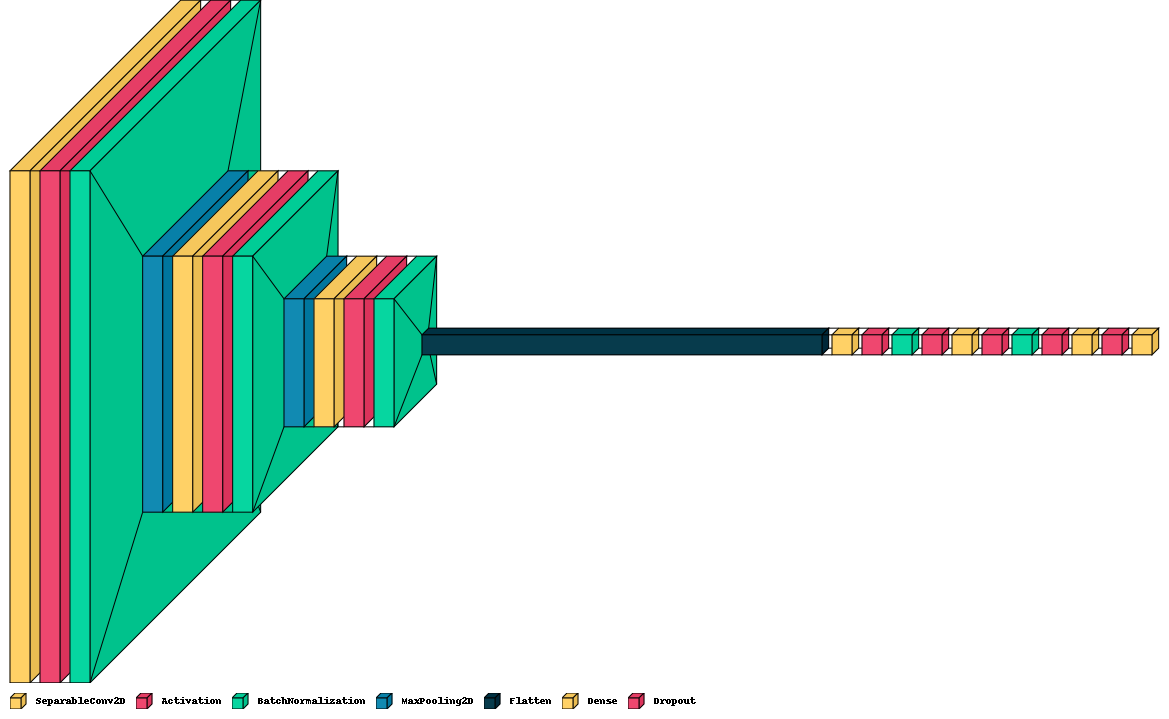}
    \caption{The Convolutional Neural Network Structure.}
    \label{fig:gate}
\end{figure}
A Convolutional Neural Network was trained on the dataset. Convolutional Neural Networks are a type of Artificial Neural Network that specializes in Image Recognition through processing the pixels of an image \cite{contributor_2018}. This is true due to its ability to identify key features that distinguish one class from another–in this case, images containing wildfires from those without them–by reducing the number of parameters for each consecutive layer. As stated by Sumit Saha, “The role of the ConvNet [Convolutional Neural Network] is to reduce the images into a form which is easier to process, without losing features which are critical for getting a good prediction” \cite{saha_2018}. Input images in Convolutional Networks used for Image Classification are passed through a filter which performs computations to condense the image's pixels and eventually foil down to a single output, which should accurately represent the input image's class. The Convolutional Neural Network used for this research was set to distinguish between images with wildfires and those without them. It was designed to take images of 128 by 128 pixels, meaning there were a total of 16384 pixels to start off with. The Convolutional Neural Network was trained on an initial learning rate of 1e-2, a batch size of 64, 15 epochs, and ended up having an accuracy of 86\%. Below is a graph of the training loss, validation loss, training accuracy, and validation accuracy.

\begin{figure}[htp]
    \centering
    \includegraphics[width=9cm]{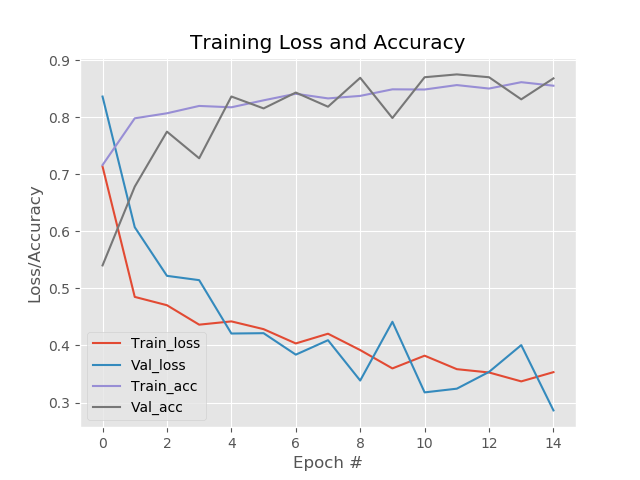}
    \caption{Training loss and validation loss drop towards 0 while training accuracy and validation accuracy trend upwards toward 1, demonstrating the model’s effectiveness through training and its ability to minimize loss and maximize accuracy.}
    \label{fig:gate}
\end{figure}

In comparison to other research performed in the field of preliminary wildfire detection, the accuracy of this Convolutional Neural Network is a little lower. However, as mentioned earlier, other research performed in this field tends to have a heavy load of bias in their datasets and use other factors that decrease the validity of the accuracy. These neighboring research papers, as mentioned in other segments of this paper, have issues regarding model robustness and generalization which often occur from the datasets presented in them that often have a bias due to the unrepresentative setting, atmosphere, lighting, etc. The dataset formed from this research mitigates and in some cases solves problems such as the ones listed above as the masks eliminate the possibility of non-wildfire-related factors interfering with the model’s weights during training. One major limitation comes in the form of a lack of data for the dataset. Although there was a total of 4000 images in the dataset (2000 images of wildfires and 2000 images of landscapes susceptible to containing wildfires, but not containing them), more images in each of the classes would have helped further increase the accuracy of the model since the features that were associated with wildfires were not very evident in the images and henceforth adding more data would have helped improve the robustness of the model. Despite this, the model still showcased an increase in accuracy and minimizing of loss as it progressed through the training procedure, which one can note as an initial progression of the Convolutional Neural Network adapting to and learning the features that determine whether or not an image contains a wildfire. Therefore, in the future, I can contact ALERTWildfire and work with the faculty on collecting more data, as that is very crucial towards the growth of this research.

\section{Conclusion}

The research for this study emphasizes the need for a more representative and less biased dataset for the purposes of preliminary wildfire detection using Convolutional Neural Networks that take in data in the form of image data. After performing background research, it became clear that there was an abundance of biased and skewed datasets in this field. Problems related to Label Imbalance and Unrepresentativeness were a common when reading through past research. However, these were often unmentioned in these studies, leading to an inaccurate accuracy and validity showcased. Several papers would often use data where the images of fires, one of the two classes for most models, would feature orientations and settings that were extremely different from what an outdoor camera would actually view in the real world. Therefore, this would result in a not very robust model as the data used for training is dissimilar to what is actually seen. Through carrying out the experimentation phase of this study, bias was mitigated from the dataset through inputting more representative images, masking parts of the image that would not declare the difference between a wildfire existing or not existing, and inputting an equal number of images with wildfires and images without them to ensure Label Imbalance was mitigated. Although there was a total of 4000 images in the dataset (2000 images of wildfires and 2000 images of landscapes susceptible to containing wildfires, but not containing them), more images in each of the classes would have helped further increase the accuracy of the model since the features that were associated with wildfires were not very evident in the images and henceforth adding more data would have helped improve the robustness of the model. Despite this, the model still showcased an increase in accuracy and minimizing of loss as it progressed through the training procedure, which one can note as an initial progression of the Convolutional Neural Network adapting to and learning the features that determine whether or not an image contains a wildfire.

\bibliographystyle{ieeetr} 

\end{document}